# FPGA-based Lane Detection System incorporating Temperature and Light Control Units

Ibrahim M. I. Qamar
Electronics and Electrical Communications Engineering
Cairo University
Giza, Egypt
ibrqamar@gmail.com

Ahmed Matar
Electronics and Electrical Communications Engineering
Cairo University
Giza, Egypt
ahmed.matar02@eng-st.cu.edu.eg

Mohamed Khaled
Electronics and Electrical Communications Engineering
Cairo University
Giza, Egypt
m7md5303@gmail.com

Saber Mahmoud
Electronics and Electrical Communications Engineering
Cairo University
Giza, Egypt
saber.mahmoud702@eng-st.cu.edu.eg

Saif Gebril
Electronics and Electrical Communications Engineering
Cairo University
Giza, Egypt
saifnasser144@gmail.com

Saleh Sharouk
Electronics and Electrical Communications Engineering
Cairo University
Giza, Egypt
salehhesham529@gmail.com

Seif M. Megahed
Electronics and Electrical Communications Engineering
Cairo University
Giza, Egypt
seifmegahed13@gmail.com

Mervat M. A. Mahmoud
Microelectronics
Electronics Research Institute
Cairo, Egypt
mervat-m@eri.sci.eg

***Abstract*— Intelligent vehicles (IVs) are one of the most important outcomes gained from the world's tendency toward automation. Applications of IVs, whether in urban roads or robot tracks, do prioritize lane (path) detection. This paper proposes an FPGA-based Lane Detector Vehicle (LDV) architecture that relies on the Sobel algorithm for edge detection. Operating on 416 x 416 images and 150 MHz, the system can generate a valid output every 1.17 ms. The valid output consists of the number of present lanes, the current lane index, as well as its right and left boundaries. Additionally, the automated light and temperature control units in the proposed system enhance its adaptability to the surrounding environmental conditions.**

## I. Introduction

The world's concern about intelligent vehicles (IV) has increased dramatically recently. Various applications were of interest, starting from just vehicle-to-vehicle communications till having almost fully automated vehicles [1]. The early implementations of lane detection systems started in the early 1980s, even before most of the modern cutting-edge technologies [1]. Recently, some existing CNN-based models have shown satisfying performance regarding this very task. However, the large number of required parameters and weights often hinders their deployment on embedded platforms [2].

This paper suggests an intelligent vehicle system featured with a hardware-friendly architecture for lane detection suitable for FPGA deployment. It is characterized by its low-resource utilization in addition to its relatively high throughput. We have organized this paper as follows: Section II highlights similar work in the literature and the reasoning behind the selected techniques. Section III explains the hardware-implemented system for lane detection. Section IV illustrates the simulation and implementation results of the hardware implementation of the lane detection system. Section V discusses the I$^2$C-interfaced supporting light and temperature control units. Section VI clarifies the conclusion and suggests new research directions for future work.

## II. Literature Review

Regardless of hardware compatibility or computational complexity, lane detection systems employed a variety of algorithms. The Canny edge-detection and Hough transform algorithms can be considered among the best candidates for performing the lane detection task. However, the complexity of the Canny algorithm stems from its multi-stage processing, double thresholding, and Gaussian filtering, making it unsuitable for hardware deployment [3]. Malmir and Shalchian in 2019 applied the Hough transform on FPGA with the addition of a backup filter (stripe detection stage) to improve the accuracy. However, the system's utilization exceeded 70K LUTs in addition to the lack of providing information about the number of lanes on the current road and the current-lane index. [4] Moreover, other simpler techniques were targeted for hardware implementation, such as Robinson, Sobel, Laplacian, and Log edge detection techniques. Humaidi et al. in 2018 use the Sobel technique, based on its lower number of convolutions and computations. Additionally, the results from the Sobel filter have proven more effective at finding edges in the images compared to other simple techniques [5]. Heuijee and Daejin in 2024 have compared the deployment of the Canny detection algorithm with the Hough transform on different boards, including FPGA and HPS. However, the final results indicated a low throughput for the system. [6]

On practical operation, various types of noise frequently affect digital images used in lane detection, typically arising from errors during the image acquisition process. This noise distorts pixel values, causing them to differ from the true intensities of the actual scene. The state-of-the-art examines several filters, including the Gaussian and the averaging filters. However, the average filter is considered to be the best candidate, giving better results than the Gaussian one [5].

This paper proposes a five-stage hardware system architecture. The system employs the Sobel technique for edge detection. After converting RGB images into grayscale, the system applies an averaging filter for noise reduction. Once that is done, post-processing represented in single-stage thresholding is performed. Finally, the lane identification block takes the system data and processes it to provide the number of lanes, the current lane index, and the boundaries on the right and left, accompanied with a valid signal for any next stage needing this information. Automated light and temperature control units were added to the system, enhancing its performance and adaptability under various environmental conditions. These supporting systems are characterized by an $I^2C$ [7] input interface to facilitate the data exchange between the system and $I^2C$-interfaced sensors.

## III. Hardware Implementation of the Lane Detection

Fig. 1 displays the proposed system block diagram. As shown, the AXI-stream protocol [8] is the system input interface, and the LKA-ACC is the unit supposed to process the system output data, where it is responsible for steering the vehicle if the motion is not centered in the current lane. All the system blocks were completely implemented using Verilog and System Verilog HDLs. First, the image is converted from RGB to grayscale, and then averaging filter removes noise. The average filter and the following-stage Sobel filter share a global FIFO (first in, first out) to write the new pixels. The lane detection and identification block, called the decision block, is responsible for identifying lane count and current-lane index in addition to its right and left boundaries. The output pixels from the Sobel filter are binarized ; if an edge is detected, then the pixel takes a high value, i.e., a white pixel, and if no edge is detected, then the pixel takes a value of zero, i.e., a black pixel. The value assignment is based on thresholding the Sobel convolutional filter output to a pre-determined value.

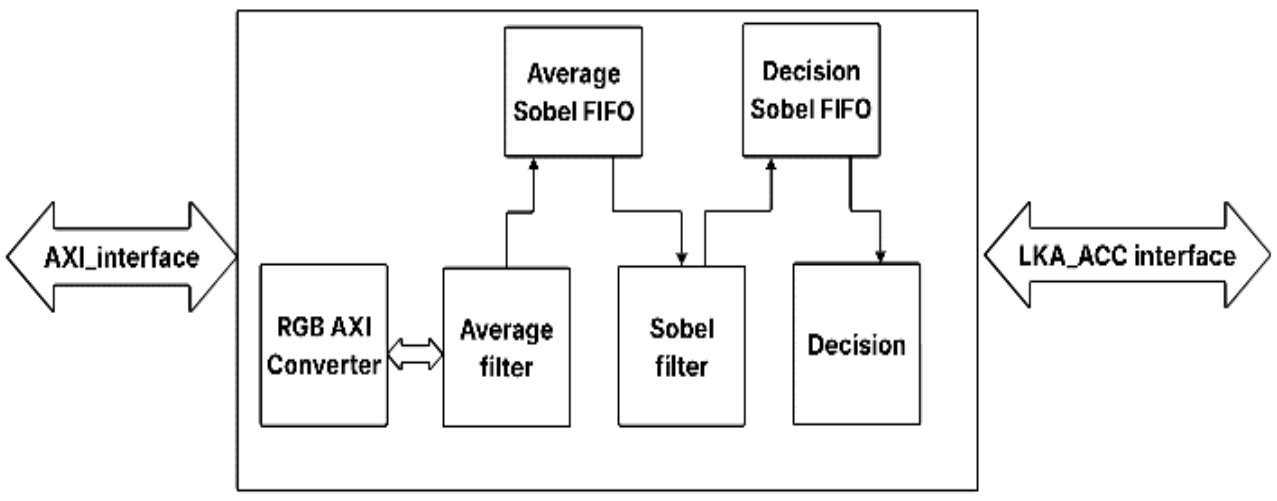

Figure 1 System Block design

TABLE 1. SLAVE AXI STREAM SIGNALS

| Input Signals | Description | No of bits |
|---|---|---|
| tdata | Input pixels to the system | 24 |
| tready | Output signal indicates start of frame | 1 |
| tvalid | Input signal indicates validity of data | 1 |

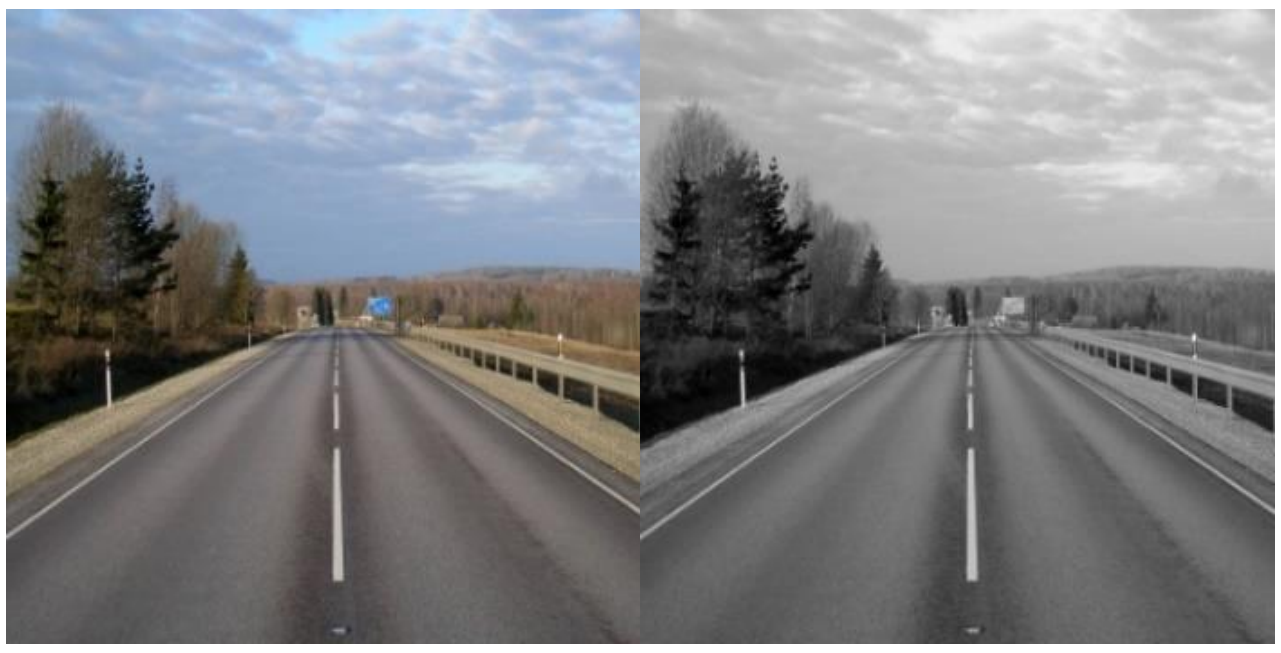

Figure 2: Original image vs. gray output from RTL

### A. Stage 1: RGB to Gray Converter

This stage is the interface between the system and the input image. The input RGB image is converted into grayscale to decrease the complexity of the computations and to facilitate the edge detection process [5]. The conversion process is expressed in Eq. 1. Moreover, this process reduces each pixel size from 24 bits (8 bits/color channel) to 8 bits.

$$Gray_{Channel} = 0.2989 * Red_{Channel} + 0.587 * Green_{Channel} + 0.114 * Blue_{Channel} \qquad (Eq. 1)$$

The block is characterized by an AXI-stream interface, which was chosen due to its high efficiency in video applications because of its signals, shown in Table 1. The block performs the conversion process using an 8-bit integer fixed-point representation to save resources and delay as well as to decrease the power needed for the conversion. Fig. 2 shows the output of the block after the RTL implementation. The results' accuracy was over 90%, which is enough for the next stage, despite the precision loss arising from the 8-bit fixed point representation.

### B. Stage 2: Noise Averaging Filtering

The averaging filter is applied to the image by sliding a 3x3 window over the whole frame with stride = 1, where the matrix weights are all, as in Eq. 2. Its primary use is to reduce the noise in the frames and minimize unwanted edges. It works as a low-pass filter. It uses the kernel described in Eq. 2 as a moving window that performs convolution on the pixels stored in the buffers. Fig. 3 shows the output from the average filter implemented using RTL. It's noted that the image has lost some of its sharpness, particularly in the edges of objects. However, this is acceptable, as the duty was to remove high-peak noise.

$$Avg\ Sliding\ Window = \begin{pmatrix} \frac{1}{9} & \frac{1}{9} & \frac{1}{9} \\ \frac{1}{9} & \frac{1}{9} & \frac{1}{9} \\ \frac{1}{9} & \frac{1}{9} & \frac{1}{9} \end{pmatrix} \qquad (Eq. 2)$$

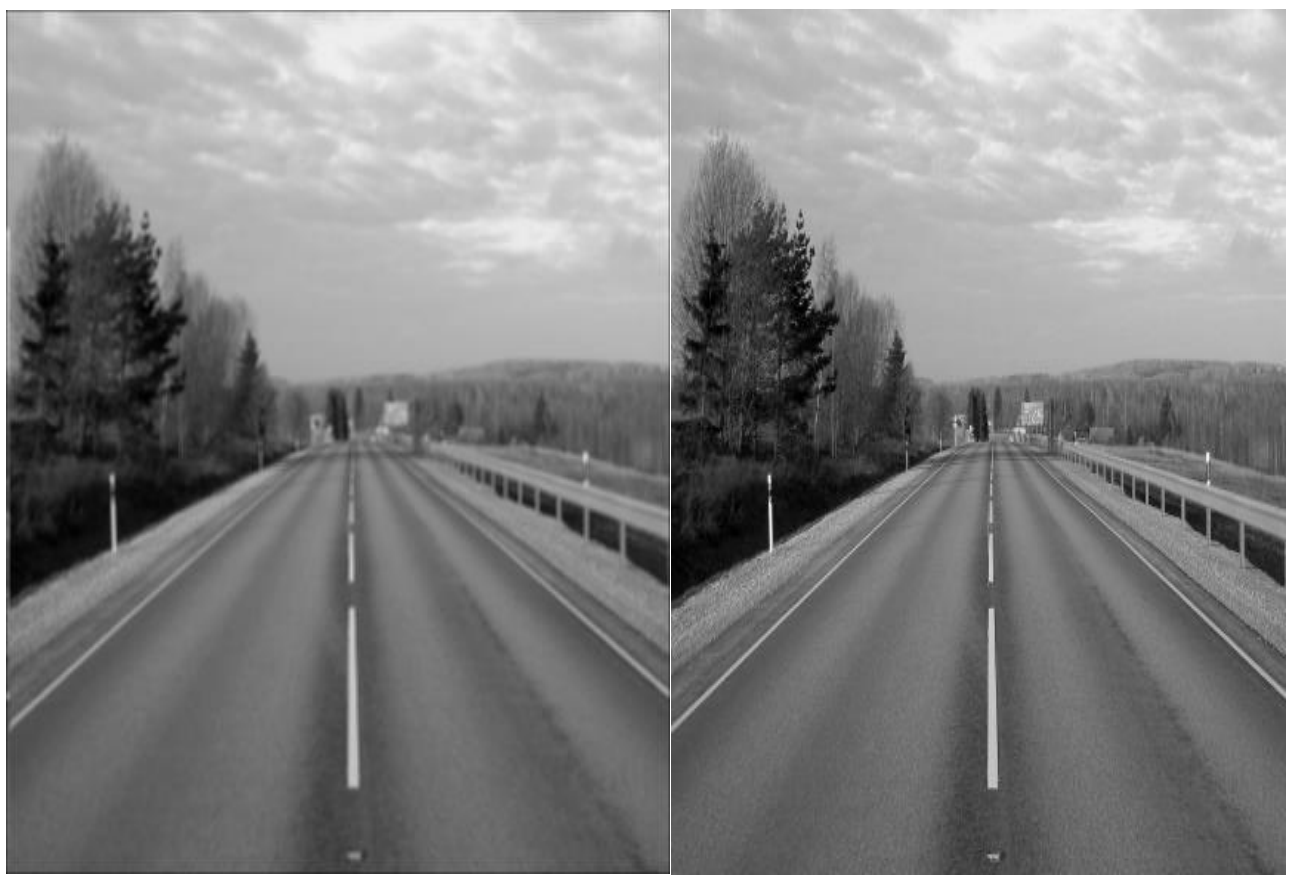

Figure 3: Avg filter output from RTL Vs Original image

*C. Stage 3: Sobel Edge Detection Technique*

The Sobel algorithm relies on two kernels (filters), Gx and Gy, where each one is responsible for being convoluted with the image in both horizontal and vertical directions, respectively. It uses two 3x3 windows for the gradients and components, as in Eqs. 3 and 4. It can be noticed that Gx is just the transpose of Gy. This operation aims to produce a value for each pixel, which we can subsequently compare to a specific threshold chosen after trials on real-case images. The stride of the convolution is 1. The filter system architecture consists of two main blocks: line buffers and the window. Line buffers read the input pixels from the interconnecting FIFO (average and Sobel interconnecting FIFO). The FIFO stores the pixels before operating on them. The Sobel filter then writes these pixels into its line buffers in order to apply the convolution process on them. Fig. 4 shows how the convolution process works with the data in the line buffers, while Fig. 5 shows the results of the Sobel filter after it has been implemented in RTL.

$$G_x \; Sliding \; Window = \begin{pmatrix} -1 & 0 & 1 \\ -2 & 0 & 2 \\ -1 & 0 & 1 \end{pmatrix} \qquad (Eq. 3)$$

$$G_y \; Sliding \; Window = \begin{pmatrix} -1 & -2 & -1 \\ 0 & 0 & 0 \\ 1 & 2 & 1 \end{pmatrix} \qquad (Eq. 4)$$

*D. Stage4: Sobel Post-Processing and Thresholding*

The gradient is calculated and compared to a threshold value to determine if the pixel is an edge or not. The output of the Sobel filter is applied to a non-maximum suppression unit. However, before that, the output of both the Gx and Gy goes into a sum of squares process to get the resulting magnitude corresponding to each pixel in the image, as in Eq. 5. To avoid complex hardware implementation, the squared gradient is used instead of getting the root of squares, and the threshold value is chosen accordingly. The produced result is compared to a pre-determined threshold, which is 22500 in the proposed implementation. Upon the magnitude comparison, each pixel in the image is now mapped to only 1 bit: '1' in case the pixel's Sobel output is more than the threshold and '0' otherwise. After simulation on real case images, the threshold value is determined through trial and error. Edge detection takes place whenever the pixel value from the Soble filter output exceeds the threshold. These binary-represented pixels are written in another global FIFO shared between Sobel and the next stage, which is the lane identification unit.

$$G_T = G_x^2 + G_y^2 \qquad (Eq. 5)$$

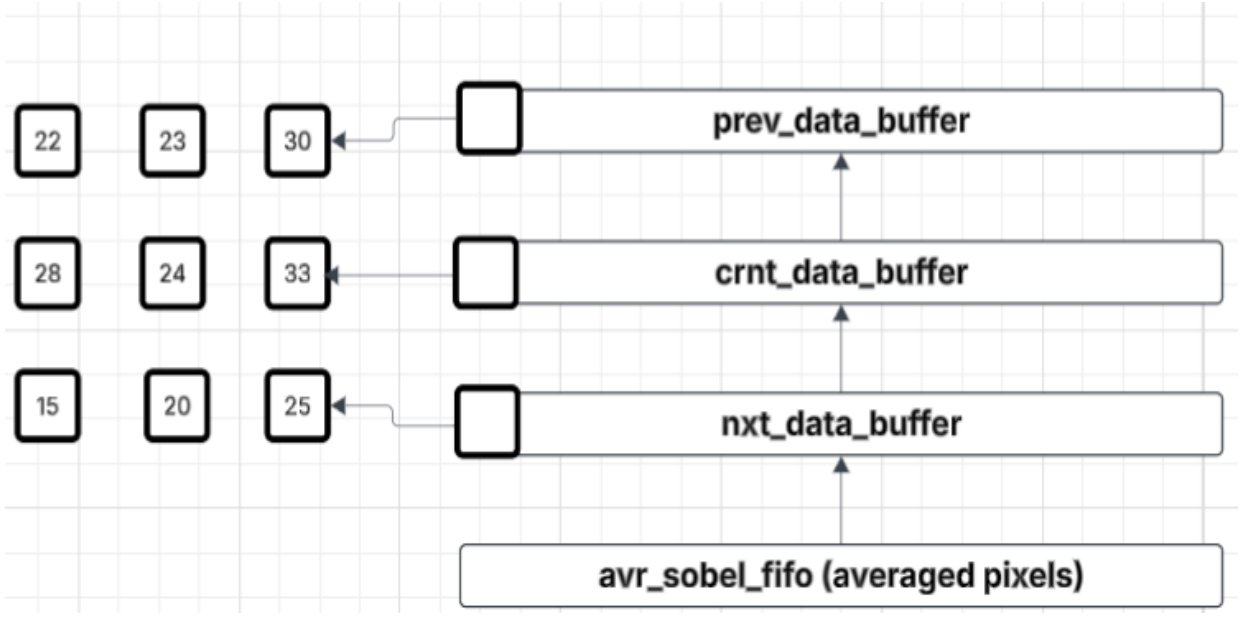

Figure 4: Proposed convolutional operation mechanism

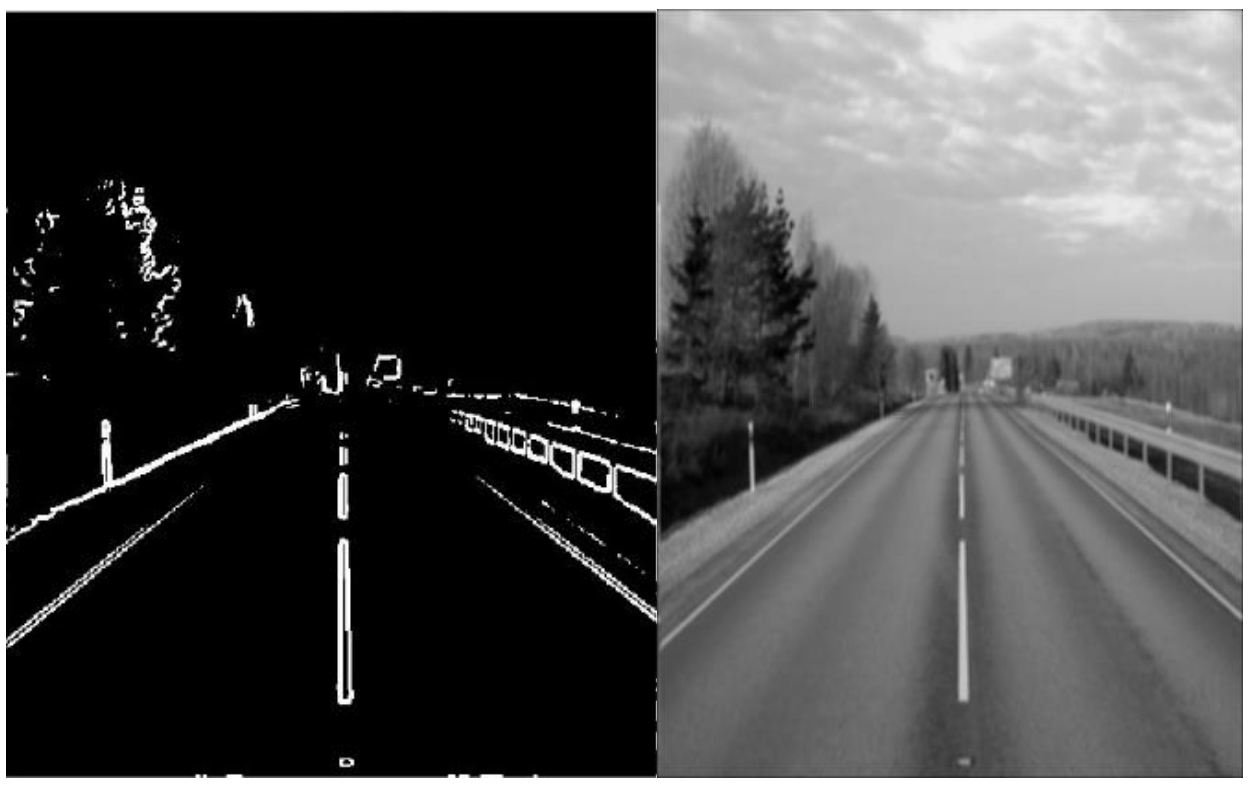

Figure 5: Binary image output from RTL vs. original image

*E. Stage 5: Lane Identification Stage*

A non-maximum suppression unit receives the output of the Sobel filter. Humans can easily identify the number of lanes and their locations in this type of image, but computers need an algorithm to do so. This algorithm does not differ much from what our brains do. The designed system reads the lines of binary pixels while keeping count of the position of the pixel in the frame. It detects clusters of edges but only counts those that have a gap between them wide enough to be a lane. This procedure is done on each line and then generates the mentioned outputs: current lane index, current lane right boundary, current lane left boundary, and the number of lanes in the road.

## IV. Simulation and Implementation Results

The system is implemented on a ZYNQ UltraScale+ ZCU102 FPGA board. The system was stimulated with images from real-life cases as well as some Python-generated images. The Python-generated images test the system's performance at a relatively large number of lanes (e.g., 13), a rare occurrence on roads. The testbench was built using System Verilog HDL. Enriching the verification environment was through utilizing the System Verilog interfaces and System Verilog Assertions. The produced coverage percentage exceeded 90%, and the system confirmed its validity with the properly preprocessed input images. Fig. 6 shows the simulation waveform.

Table 2 displays how much of the FPGA resources was utilized, showing that the proposed lane detection system uses less of the board compared to other results with different algorithms. The system operates on 150 MHz, allowing for a valid output every 1.17 ms for a single frame, achieving a magnificent rate of processing. The latency is represented by counting the clock cycles in each stage, then add them together to calculate the total latency. The following calculations are only for the critical stages of the system: RGB to gray converter, average, and Sobel filters. Table 3 shows the latency for each of them.

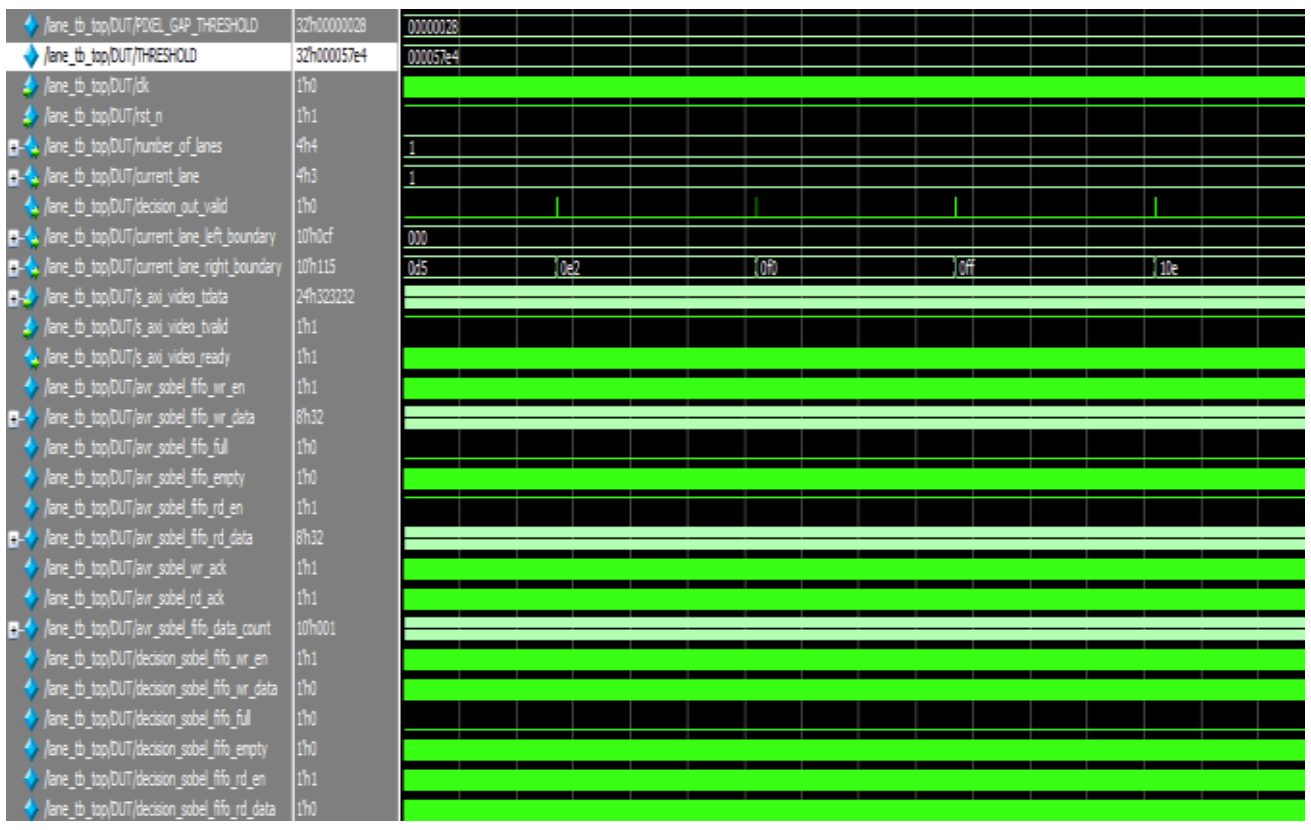

Figure 6: System Simulation resulted waveform

TABLE2: FPGA UTILIZATION

| Resource | Utilization | Utilization % |
|---|---|---|
| LUT | 11551 | 4.21 |
| FF | 25205 | 4.6 |
| DSP | 14 | 0.56 |
| BUFG | 1 | 0.25 |

TABLE 3. LATENCY OF THE SYSTEM

| **Block** | **Latency (Clock cycles)** |
|---|---|
| RGB to Gray convetor | 1 |
| Average Filter | 422 |
| Sobel Filter | 422 |

The RGB to gray takes 1 clock cycle. It receives the data and performs the conversion operation as stated in Eq. 1. The average filter latency isn't effective unless the system has just started and the line buffers are empty. When Line Buffer 1 is empty, the pixels are directly written into it, ignoring Line Buffer 2. Then the buffer is filled in N clock cycles, where N is the number of pixels in the row of the frame, and then it waits for another 3 cycles for 3 pixels to be written inside Line Buffer 2. Thereafter, the pixels are written into the shift register, which takes 2 clock cycles to form the window and 1 clock cycle to calculate the convolution. So, the total number of cycles is N + 6. In our system, N = 416. The Sobel Filter has the same number of cycles as the Average Filter, which takes 422 cycles. We calculate the throughput as the output per clock cycle. We should analyze each stage independently to calculate the throughput.

For the RGB to Gray Converter, its latency is 1 clock cycle, and the throughput is 1 pixel/clock cycle if the Average Filter is ready to take data in. The latency is 422 clock cycles for the first pixel. Afterward, the block produces one output pixel per clock cycle. For a whole frame of 416 x 416, the throughput is calculated by $Throughput = \frac{416^2}{422+416^2} \approx 1$. The pipeline design mitigates the impact of large latency, making it negligible in the system. Once the first row is loaded into the buffers of either the average or the Sobel, the output from the filters is always 1 pixel/clock cycle.

## V. I2C-BASED SUPPORTING SYSTEMS

The proposed system was enhanced with automated light and temperature control units to increase its adaptability to outer environments as well as improve the luxury of the cruise in case it was utilized on urban roads for personal cars. Figs. 7 and 8 show these control units' architectures. They are characterized by their I²C interface. Moreover, the resolution of the output enhances the power saving since the light or the AC isn't always at full power in their ON cases but at a reasonable adjustable level.

### A. *Light Control Unit:*

The light control unit architecture contains a clock divider, an I²C controller, and a comparator. The clock divider module reduces the system clock frequency to generate a slower clock signal used by the I²C controller. This feature is necessary to meet the timing requirements of standard I²C communication. We parameterize the divider to enable flexible output frequency adjustment based on the system clock. The I²C controller is responsible for communicating with the digital light sensor. It first initializes the sensor by writing to its configuration registers, then periodically reads brightness data. The module handles start and stop conditions, addressing, and data transfer sequences and asserts a valid signal once new sensor data is available. When the I/C controller asserts a valid signal, the comparator module receives 16-bit brightness data. It linearly scales the data down to 12-bit resolution for DAC compatibility. The scaled value is then compared against a predefined threshold. If the value

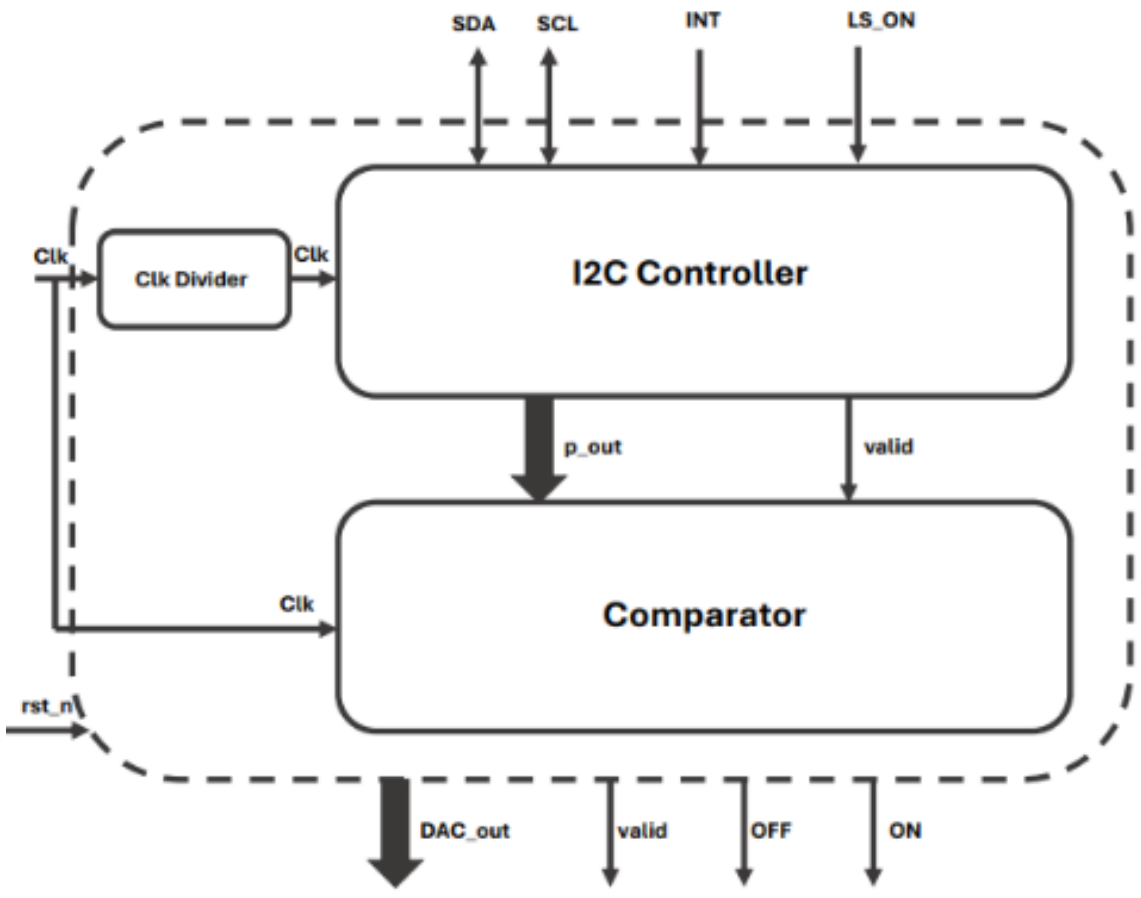

Figure 7: Light Control Unit Architecture

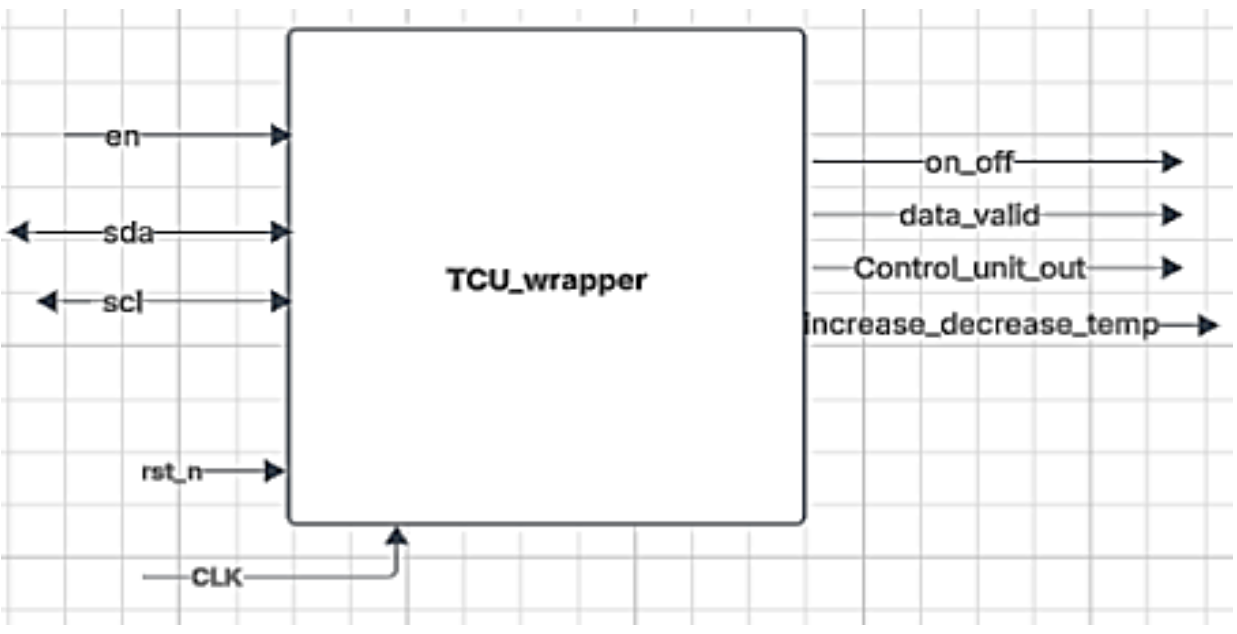

Figure 8: Temperature Control Unit Architecture

is below the threshold, the module signals to the next stage module to open the light.

### B. Temperature Control Unit:

Environmental temperature regulation is a critical aspect of modern smart control systems. The architecture of the temperature control unit includes an I²C controller. This module handles communication with the digital temperature sensor over the I²C protocol. It initiates read commands, accesses the temperature register, and retrieves two 8-bit data bytes representing the measured temperature.

The Temperature Control Unit (TCU) is designed to manage temperature-sensitive systems such as air conditioners. It receives raw temperature data from the I²C controller and converts it to degrees Celsius using a dedicated temperature calculation module. The TCU continuously monitors the temperature and compares it to a reference value of 25°C. If the deviation exceeds a predefined noise threshold, it generates a proportional control signal through a DAC output. Based on this deviation, the module determines whether the air conditioner should be activated and adjusts the output power accordingly. The greater the deviation from 25°C, the higher the required cooling or heating effort. Eq. 6 demonstrates the temperature calculation in Celsius.

$$\text{Temperature} = \text{sensor_out} * \text{resolution_of_sensor} \qquad (Eq.6)$$

## VI. Conclusion and Future work

The proposed system introduces a lane detector vehicle with low utilization, despite its high capability in different circumstances. Moreover, it is enhanced with automated light and temperature control units in order to adapt to different operating environmental conditions. The light control unit has also removed the need for a night vision camera due to the feature added of adjusting the light for any normal camera to be able to analyze the track. However, the system can be enhanced in future versions with an increase in its throughput by forcing the decision block to generate output earlier as well as using more than one filter to employ parallelism, which indeed can improve the system's overall performance.

## Acknowldgements

We want to show our gratitude towards Dr. Mohsen Mahroos (Deceased) , the professor at Cairo University, EECE Department for his support and mentorship during the project progression.